%% file: main-tgrs.tex
\documentclass[twocolumn]{IEEEtran}

\usepackage[pdftex]{graphicx} 
\usepackage{amsmath} 
\usepackage{algorithmic}
\usepackage{array} 
\usepackage[caption=false,font=footnotesize]{subfig} 
\usepackage{dblfloatfix}
\usepackage{url} 

\usepackage{microtype}
\usepackage[utf8]{inputenc} 
\usepackage[T1]{fontenc}    
\usepackage{paralist} 
\usepackage{marvosym} 
\usepackage{bm} 
\usepackage{bbm} 
\usepackage{amssymb}   
\usepackage{mathrsfs} 
\usepackage{tabularray} 
\UseTblrLibrary{booktabs}       
\usepackage{tabularx} 
\usepackage[flushleft]{threeparttable} 
\usepackage{colortbl} 
\usepackage{makecell} 
\usepackage{multirow} 
\usepackage{nicefrac} 
\usepackage[dvipsnames]{xcolor} 
\usepackage{soul} 
\setul{0.5ex}{0.3ex}

\definecolor{citecolor}{RGB}{34,139,34}
\usepackage[pagebackref=true,breaklinks=true,letterpaper=true,colorlinks,
    citecolor=citecolor,bookmarks=false]{hyperref}
\usepackage{cleveref}  
\crefformat{figure}{Fig.~#2#1#3}
\crefformat{equation}{Eq.~(#2#1#3)}
\crefformat{table}{Table~#2#1#3}
\crefformat{section}{Section~#2#1#3}
\crefformat{algorithm}{Algorithm~#2#1#3}
\crefformat{appendix}{Appendix #2#1#3}
\crefformat{subsection}{subsection~#2#1#3}
\usepackage{cellspace}
\usepackage[numbers,sort&compress]{natbib} 

\usepackage{xspace} 

\usepackage{pifont} 
\definecolor{Gray}{rgb}{0.9,0.9,0.9}
\definecolor{LightCyan}{rgb}{0.88,1,1}
\newcolumntype{a}{>{\columncolor{Gray}}c}
\newcolumntype{b}{>{\columncolor{white}}c}

\begin{document}

\setlength{\abovedisplayskip}{.5\baselineskip} 
\setlength{\belowdisplayskip}{.5\baselineskip} 

\title{AuxDet: Auxiliary Metadata Matters for Omni-Domain Infrared Small Target Detection}

\input{journal-contents/1-authors.tex}

\maketitle

\input{journal-contents/2-abstract.tex}
\input{journal-contents/3-introduction.tex}

\input{journal-contents/4-related-work.tex}
\input{journal-contents/5-method.tex}
\input{journal-contents/6-experiment.tex}

\input{journal-contents/7-analysis.tex}
\input{journal-contents/8-conclusion.tex}

\bibliographystyle{IEEEtran}
\bibliography{./reference.bib}

\end{document}

%% file: journal-contents/1-authors.tex

\author{
  Yangting~Shi,
  Yinfei~Zhu,
  Renjie~He,
  Le~Hui,
  Meng~Cai,
    Ming-Ming~Cheng,
  Yimian~Dai
  \thanks{
    This work was supported by
      National Key Research and Development Program of China (2023YFF0906203), 
      the National Natural Science Foundation of China (62301261, 
        62306238) 
        and
      the Fundamental Research Funds for the Central Universities  
    \emph{The first two authors contributed equally to this work. (Corresponding author: Renjie He and Yimian Dai).}
    }


  \thanks{
    Yangting Shi, Renjie He and Le Hui are with the School of Electronics and Information at Northwestern Polytechnical University.
    Le Hui also holds a position at the Shaanxi Key Laboratory of Information Acquisition and Processing.
    (e-mail:
    \href{mailto:yangting.shih@gmail.com}{yangting.shih@gmail.com};
    \href{mailto:davidhrj@nwpu.edu.cn}{davidhrj@nwpu.edu.cn})
    }

  \thanks{
    Yinfei Zhu and Meng Cai are with the Luoyang Institute of Electro-Optical Equipment under AVIC (Aviation Industry Corporation of China).
    \href{mailto:zhuyf\_613@163.com}{zhuyf\_613@163.com}.
  }

    \thanks{
     Ming-Ming Cheng and Yimian Dai are with VCIP, College of Computer Science, Nankai University. 
  (e-mail:
  \href{mailto:yimian.dai@gmail.com}{yimian.dai@gmail.com})
  \href{mailto:cmm@nankai.edu.cn}{cmm@nankai.edu.cn};)
  }

}

%% file: journal-contents/2-abstract.tex

\begin{abstract}

Omni-domain infrared small target detection (Omni-IRSTD) poses formidable challenges, as a single model must seamlessly adapt to diverse imaging systems, varying resolutions, and multiple spectral bands simultaneously. Current approaches predominantly rely on visual-only modeling paradigms that not only struggle with complex background interference and inherently scarce target features, but also exhibit limited generalization capabilities across complex omni-scene environments where significant domain shifts and appearance variations occur. In this work, we reveal a critical oversight in existing paradigms: the neglect of readily available auxiliary metadata describing imaging parameters and acquisition conditions, such as spectral bands, sensor platforms, resolution, and observation perspectives. To address this limitation, we propose the Auxiliary Metadata Driven Infrared Small Target Detector (AuxDet), a novel multimodal framework that is the first to incorporate metadata into the IRSTD paradigm for scene-aware optimization. Through a high-dimensional fusion module based on multi-layer perceptrons (MLPs), AuxDet dynamically integrates metadata semantics with visual features, guiding adaptive representation learning for each individual sample. Additionally, we design a lightweight prior-initialized enhancement module using 1D convolutional blocks to further refine fused features and recover fine-grained target cues. Extensive experiments on the challenging WideIRSTD-Full benchmark demonstrate that AuxDet consistently outperforms state-of-the-art methods, validating the critical role of auxiliary information in improving robustness and accuracy in omni-domain IRSTD tasks. 
Code is available at \href{https://github.com/GrokCV/AuxDet}{https://github.com/GrokCV/AuxDet}.
\end{abstract}

\vspace{-1\baselineskip}

%% file: journal-contents/3-introduction.tex

\section{Introduction} \label{sec:introduction}


\IEEEPARstart{I}{nfrared} imaging offers unparalleled advantages in capturing  target information under adverse weather conditions and limited visibility scenarios, which has enabled wide-ranging applications in marine resource utilization~\cite{teutsch2010classification}, precision navigation~\cite{wang2020robust}, and ecological monitoring~\cite{hao2024infrared}.
However, due to the long-range imaging distance, targets of interest typically exhibit extremely limited spatial resolution, lack texture and structure, and often blend into complex backgrounds~\cite{kou2023infrared}, making them easily overwhelmed by environmental clutter.
Therefore, infrared small target detection (IRSTD) still remains a  challenging task for computer vision.

\begin{figure}[t]
	\vspace{0pt}
	\begin{center}
		\setlength{\fboxrule}{0pt}
        \vbox{\includegraphics[width=\columnwidth]{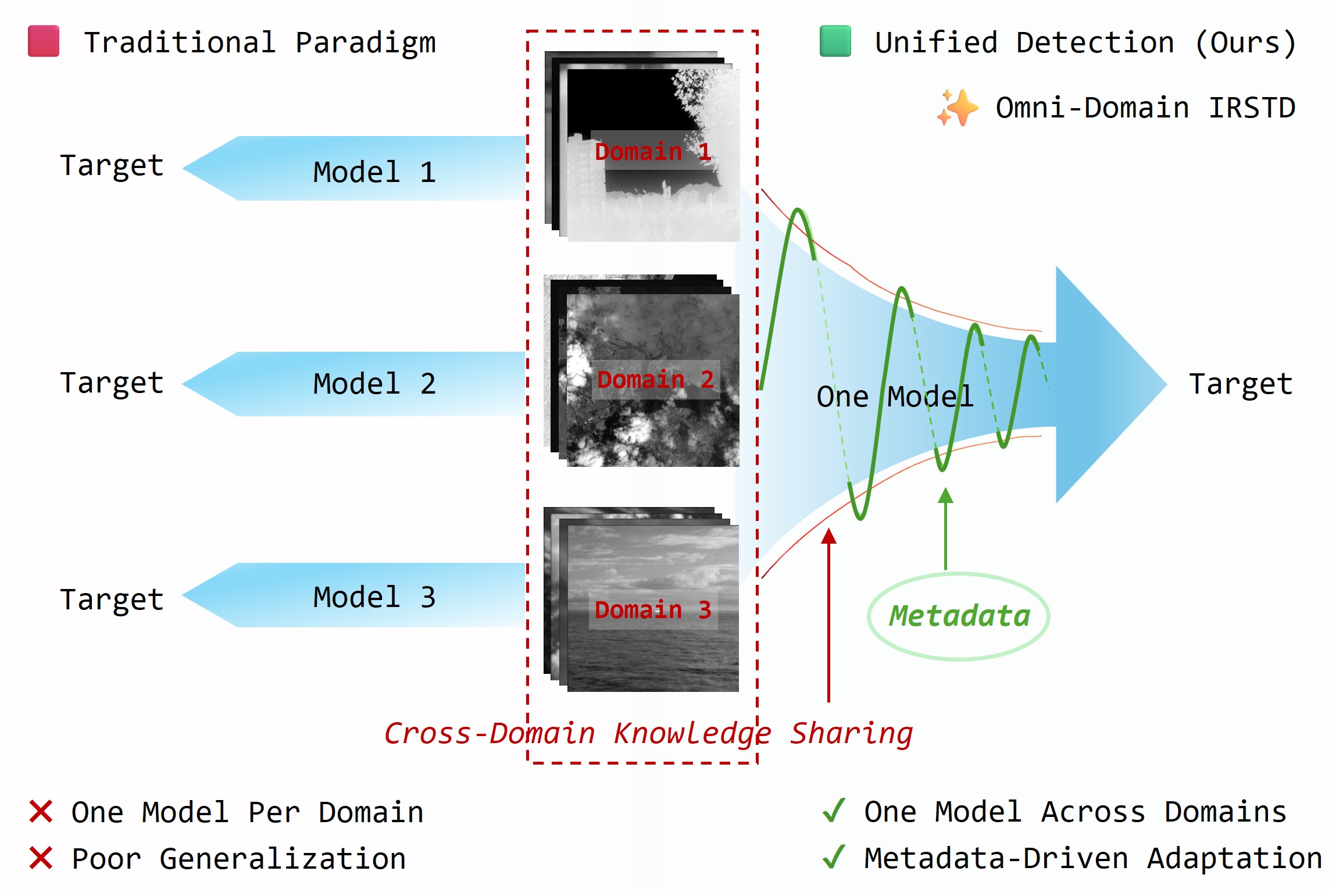}}
	\end{center}	
	\vspace{-12pt}
	\caption{
    Traditional vs. unified detection. AuxDet leverages metadata for effective omni-domain generalization using one unified model.
    }
	\label{fig:motivation}
	\vspace{-6pt}
\end{figure}

One of the major difficulties in infrared small target detection lies in the extremely small target size and the lack of distinguishable textures or colors, making them prone to being submerged in background noise. Traditional detection methods rely on prior assumptions about targets and backgrounds, using mathematical models for extraction~\cite{2006The, bai2010analysis, wei2016multiscale, kim2012scale, deshpande1999max, bae2011small, wang2006small, TIP13IPI,dai2016infrared}. However, as background complexity increases, these methods struggle with robustness and adaptability.

Recently, with the rise of deep learning, many efforts have attempted to address IRSTD through neural architectures that emphasize robust feature extraction and refined attention mechanisms~\cite{chi2025contrast}.
Representative works like ALCNet~\cite{dai2021attentional} introduced attention-guided local contrast mechanisms to amplify subtle target-background discrepancies, while ISNet~\cite{zhang2022isnet} developed infrared-specific spatial transformers for noise-robust feature learning. 
Despite these advances, existing methods exhibit two critical shortcomings when scaling to real-world scenarios:
\begin{itemize}
    \item \textbf{Over-Specialization}: 
    Most approaches optimize for narrow domains, such as land-based Long-Wave Infrared (LWIR) or space-based Short-Wave Infrared (SWIR) imagery, failing to generalize across the sensor-platform-spectrum continuum of omni-domain applications.
    \item \textbf{Over-Engineering}: 
    Researchers increasingly rely on domain-specific techniques to enhance detection accuracy, such as pre-identifying sea regions through background decomposition for maritime targets~\cite{hu2024smpisd}. While these methods reduce task complexity in narrow domains, they inherently compromise cross-domain generalization capabilities.
\end{itemize}

\begin{figure}[t]
	\vspace{0pt}
	\begin{center}
		\setlength{\fboxrule}{0pt}
		\vbox{\includegraphics[width=\columnwidth]{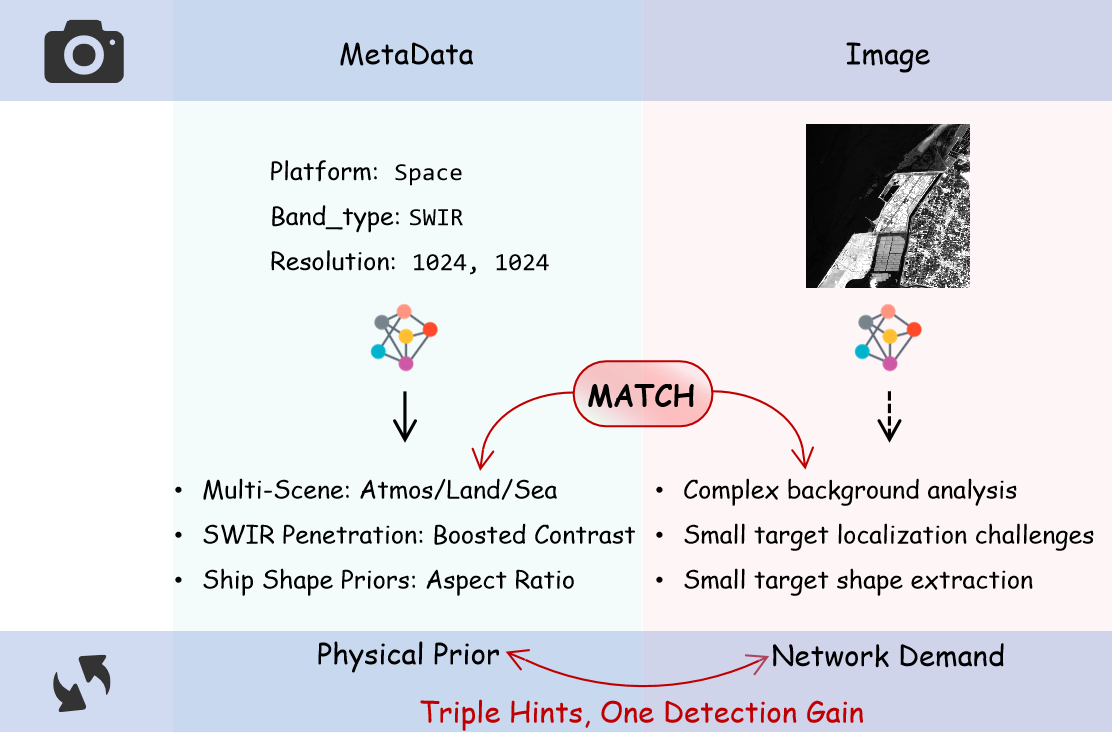}}
	\end{center}	
	\vspace{-12pt}
	\caption{
    The benefit of leveraging auxiliary information to assist infrared small target detection lies in its ability to guide the network towards inference objectives that typically require deep architectures, but at an extremely low computational cost.
    }
	\label{fig:motiv-meta}
	\vspace{-6pt}
\end{figure}

\begin{figure*}[hbtp]
\vspace{-1\baselineskip}
  \centering
  \includegraphics[width=0.98\linewidth]{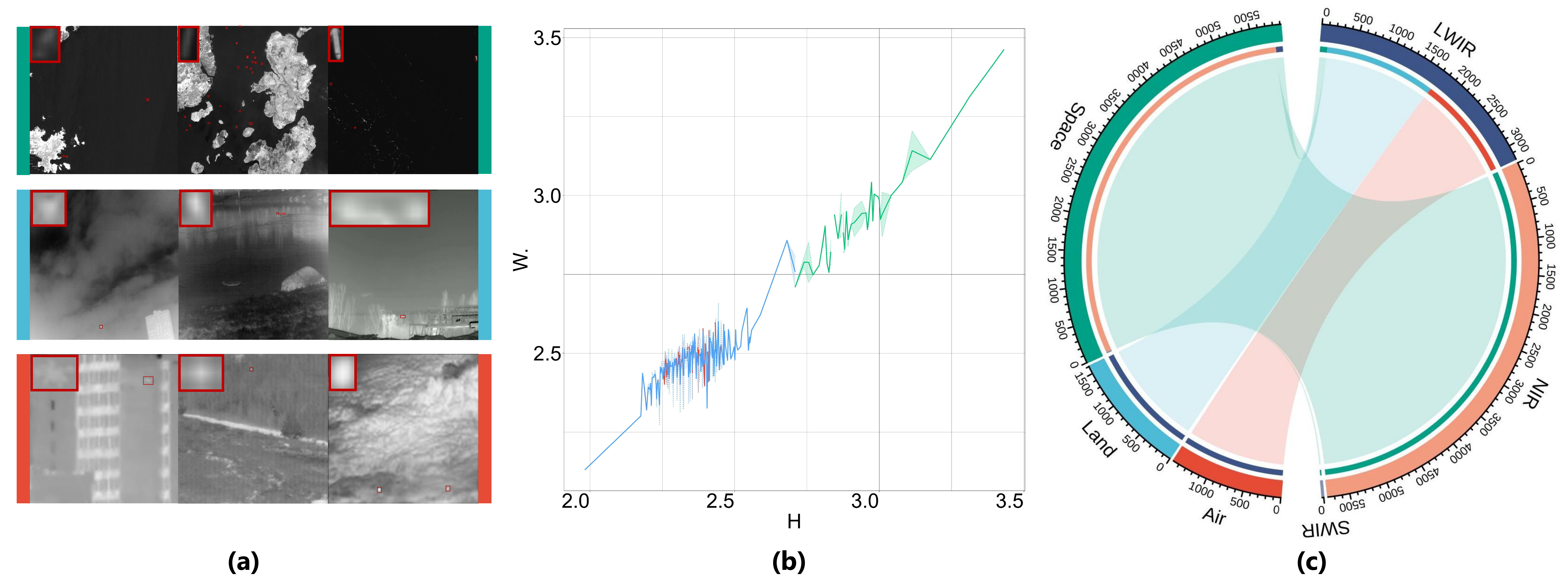}
  \vspace{-0.5\baselineskip}
  \caption{Multidimensional distribution characteristics of the WideIRSTD-Full Dataset. (a) \textbf{Cross-platform target-background contrast}: Resolution-normalized visualization of platform-specific heterogeneity in background complexity and small target morphology. (b) \textbf{Cross-platform resolution distribution}: x/y-axes represent log-transformed image height and width ranges, revealing resolution disparities across observation platforms (Air/Land/Space). (c) \textbf{Platform-band correlation chord diagram}: Coupling relationships between observation platforms (Air/Land/Space) and infrared bands (Long-Wave Infrared, LWIR; Near-Infrared, NIR; Short-Wave Infrared, SWIR). Key findings include: (1) \textit{Inter-platform resolution intersection}: Significant overlapping regions between Air (green) and Land (blue) platforms necessitate cross-platform multi-scale detection; (2) \textit{Platform-dependent target characteristics}: Distinct platform-specific distributions of target resolutions and categories induce varying detection difficulties; (3) \textit{Spectral-platform association rules}: Space-based observations (orange) are NIR-dominant, while aerial/terrestrial platforms (green/blue) solely utilize LWIR.
  }
  \label{fig:Dataset}
\vspace{-1\baselineskip}
\end{figure*}

To meet the growing demand for large-scale, multi-platform, and real-world deployable detection systems, in this paper, we formalize \textbf{a new task}, \textbf{Omni-Domain Infrared Small Target Detection} (\textbf{Omni-IRSTD}).
As shown in Fig.~\ref{fig:motivation} and Fig.~\ref{fig:Dataset}, this task aims to unify the IRSTD paradigm across heterogeneous sensing domains by addressing four core dimensions: (1) multi-platform observations (e.g., land-based, aerial-based, space-based), (2) multi-target types (point-like, spot-like, extended targets), (3) multi-spectral bands (e.g., LWIR, Near-Infrared (NIR), SWIR), and (4) multi-resolution imaging (from $256\times 256$ to $6000 \times 6000$)~\cite{Li2024First}.
Compared to traditional specific-domain IRSTD, Omni-IRSTD presents substantial challenges due to dramatic variations in environmental contexts, background characteristics, and target appearance distributions.
These cross-domain disparities significantly impact model generalization capabilities, as evidenced in Tab.~\ref{tab:WideIRSTD_comparison}, where state-of-the-art (SOTA) methods experience catastrophic performance degradation on omni-domain benchmarks.
This performance gap underscores the need for a paradigm shift to address the challenges posed by the diverse and complex nature of Omni-IRSTD.

We argue that the fundamental limitation of existing approaches stems from their paradoxical pursuit of unified feature representation across physically divergent imaging domains.
When naively aggregating multi-domain training data, the gradient directions from heterogeneous domains exhibit severe conflicts in parameter space forcing the network into suboptimal equilibrium states that compromise all domains~\cite{kou2023infrared}.
This observation sparks a natural insight: since the ideal feature representations differ across domains is neither practical nor effective, instead, the model should be aware of which domain a given input belongs to, and adapt its representation strategy accordingly.
However, conventional single-modality visual models lack the capacity to infer domain-specific imaging conditions solely from raw pixel data~\cite{zhang2025mirsam}.
This raises a critical question: \textit{can a model know the domain of an input sample in advance to enable such adaptability?}


Our answer is yes.
In practical infrared sensing systems, each captured image is inherently accompanied by a set of auxiliary metadata that records the physical conditions under which the image was acquired, such as sensor platform, spectral band, resolution, and observation angle. 
Rather than being incidental or redundant, these metadata descriptors serve as compact, semantic-rich representations of the domain to which each sample belongs~\cite{bernhard2024context}. 
In essence, \textbf{metadata offers a low-dimensional yet highly informative abstraction of the imaging domain}, encoding the very factors that drive domain-specific feature distributions and appearance patterns. 
By explicitly incorporating this metadata into the learning process, as shown in Fig.~\ref{fig:motivation}, we can transform the detection paradigm from a passive, vision-only modeling regime to an active, context-aware framework, where the model is empowered to adapt its representation strategy conditioned on the domain characteristics of each input sample~\cite{gao2025gasc}. 


More concretely, these metadata elements encode domain-specific priors through three key dimensions, as illustrated in Fig.~\ref{fig:motiv-meta}:
\begin{itemize}
    \item \textbf{Observation platform (e.g., air-based, space-based, or land-based)} 
    inherently conveys background complexity and target distribution characteristics through platform-specific attributes (e.g., in space-based remote sensing, the probability of certain geometric features in maritime targets increases significantly). At the same time, this information implicitly encodes atmospheric transmittance, aiding in the estimation of true radiative intensity. Consequently, during target detection, it guides the model to adopt more precise and adaptive processing strategies based on the observation source.
    
    \item \textbf{Spectral band} 
    of image acquisition(e.g., LWIR, SWIR, or NIR) serves as a distinctive signature of electromagnetic radiation. Variations in physical mechanisms governing different wavelength reflections directly determine the formation of target-background contrast and the statistical properties of noise distribution.
    
    \item \textbf{Image resolution} 
    acts as a crucial prior, enabling the model to infer the actual target size based on scaling factors, thus providing more refined decision support for detection tasks.
\end{itemize}

The combination of just a few characters representing these metadata elements offers highly valuable "differentiation descriptors" and "methodological guidance" for both images and networks, thereby circumventing the need for complex model design and structural optimizations in omni-domain infrared small target detection scenarios.



Building upon these insights, we propose the Auxiliary Metadata Driven Detector (AuxDet), a novel framework for Omni-IRSTD that redefines the conventional visual-only paradigm by leveraging auxiliary metadata for multimodal scene-aware optimization.
Unlike existing methods, AuxDet dynamically integrates metadata, such as spectral bands, sensor parameters, and observation conditions, with visual features through a carefully designed Multi-Modal Dynamic Modulation (M2DM) module. This module employs a multi-layer perceptron (MLP)-based fusion strategy to embed metadata semantics into the feature space, enabling sample-specific modulation that enhances target discriminability amid complex backgrounds.
By incorporating cross-layer differential features, M2DM further compensates for the scarcity of small target cues, fostering robust contextual representations. 
Further, to address the trade-off between high-level semantics and fine-grained target details, we propose a Lightweight Edge Enhancement Module (LEEM), which utilizes 1D convolutional blocks initialized with local differential priors to refine edge responses and boost detection precision.

We summarize the main contributions of the paper as follows: 
\begin{itemize}
\item We are the first to introduce metadata into the IRSTD task, transforming the traditional vision-only detection paradigm into a multimodal, scene-aware framework.

\item We propose the M2DM module, which injects auxiliary metadata into both channel-wise and spatial modulation processes. This design supports sample-specific adaptation of feature representations, empowering the detector with dynamic adaptability across complex and diverse environments.

\item We design the LEEM, an ultra-lightweight module that amplifies edge responses essential for detecting extremely small targets.
\end{itemize}

Extensive experiments on the challenging WideIRSTD-Full benchmark~\cite{Li2024First} demonstrate the effectiveness of our approach. 
Notably, compared to visual-only baselines, AuxDet yields over 30\% improvement in detection accuracy, highlighting the critical role of metadata in enabling robust and generalizable omni-domain IRSTD.






%% file: journal-contents/4-related-work.tex

\section{Related Work} \label{sec:related}


\subsection{Infrared Small Target Detection}

Traditional infrared small target detection methods mainly rely on the characteristics of targets and backgrounds for target extraction. Depending on the prior knowledge or theoretical basis, these methods can be classified into: 1) methods based on local information, such as morphological Top-Hat transforms~\cite{2006The, bai2010analysis} and local contrast metrics inspired by the human visual system~\cite{wei2016multiscale, kim2012scale}. However, the characteristics of small targets make it difficult to match traditional structural elements, and the local contrast limits the utilization of global information, rendering these methods unable to accurately differentiate the subtle differences between targets and backgrounds. 2) Filtering-based methods, including spatial domain~\cite{deshpande1999max, bae2011small} and frequency domain~\cite{wang2006small} filtering techniques, which fail to effectively separate targets from backgrounds when the spatial characteristics of the background and target are similar, and the forward and reverse transformations require considerable computational resources. 3) Low-rank representation-based methods, such as the Infrared Patch-Imag model~\cite{TIP13IPI} and Reweighted Infrared Patch-Tensor model~\cite{dai2016infrared}, which assume that the background image exhibits low-rank characteristics when high-frequency components, such as random noise, are sparse. However, in practical applications, infrared image backgrounds are often more complex, and the low-rank assumption does not hold.
While effective in controlled settings, these approaches lack generalization across diverse imaging scenarios where background variability and noise overwhelm their simplistic priors.


With the rapid development of deep learning and the release of various public datasets, deep learning-based detection methods have gradually emerged as the mainstream. These methods leverage the powerful modeling ability of deep neural networks to automatically learn features related to infrared small targets.  
Existing methods mainly focus on exploring network architectures, utilizing contrastive information, and emphasizing edge and shape information~\cite{zhao2024multi}.


In terms of network architecture, Dai et al.~\cite{dai2021asymmetric} proposed an asymmetric context modulation mechanism to effectively solve the issue of scale mismatch between infrared small targets and objects in general datasets by employing complementary top-down and bottom-up modulation paths. 
Wu et al.~\cite{wu2022uiu} proposed an interactive cross-attention nested U-Net network to achieve multi-level target representation learning.  Dai et al.~\cite{dai2023one} iteratively refined target bounding boxes through top-down cascaded detection heads. To establish long-range dependencies, 
Pan et al.~\cite{pan2023abc} 
and Yuan et al.~\cite{yuan2024sctransnet} introduced the visual Transformer architecture into infrared weak small target detection. 
Zhang et al.~\cite{zhang2025irsam} introduced the Arbitrary Segmentation Everything model into the IRSTD domain, demonstrating outstanding performance.

With reference to contrastive information utilization, Dai et al.~\cite{dai2021attentional} integrated local contrast priors into convolutional networks to make full use of domain knowledge. Yu et al.~\cite{yu2022pay} introduced a local contrast learning structure, inspired by non-deep learning methods, reducing the dependence on dataset sample size. 
Hu et al.~\cite{hu2024smpisd} used a two-layer sliding window structure in their scene semantic extractor, using a larger window to compute background pixel changes and a smaller window to capture local contrast for the target, minimizing false alarms and assessing the likelihood of target appearance. Wang et al.~\cite{wang2025paying} fused local contrast enhancement knowledge into the convolution operator design, effectively mitigating the blind nature and local extrema problem associated with autonomous learning in deep neural networks.

Regarding edge and shape information, Zhang et al.~\cite{zhang2022isnet} proposed a new infrared shape network (ISNet), designed with edge blocks inspired by Taylor’s finite difference to extract useful edge features for predicting the precise shape of targets. Lin et al.~\cite{lin2024learning} used central difference convolution and large kernel convolutions to extract discriminative representations of infrared small targets. Shi et al.~\cite{shi2024hs} argued that the details and edges of tiny objects correspond to high-frequency components in the image spectrum and incorporated a high-frequency perception module in the lateral connections of the feature pyramid to generate high-frequency responses, enriching small target information in the backbone network features. 

However, these models are trained on datasets of specific application scenarios~\cite{zhang2022isnet, dai2021asymmetric}, which suffer catastrophic performance drops when deployed to unseen domains due to domain-specific feature overfitting.
Moreover, omni-domain IRSTD task amplifies this critical limitation, demanding a single model to perform reliably across multiple imaging systems and conditions.

In contrast to previous works, our work introduces a fundamental shift in both task formulation and modeling paradigm. We position AuxDet as the first framework to explicitly tackle omni-domain IRSTD through metadata-guided multimodal learning:
\begin{enumerate}
    \item The Omni-IRSTD task formulation departs from conventional single-domain IRSTD settings by addressing the challenge of cross-domain generalization. To the best of our knowledge, this is the first work to explicitly emphasize and benchmark cross-domain generalization as a core objective in the IRSTD field.
    \item Our modeling paradigm of AuxDet moves beyond the traditional vision-only detection framework by incorporating auxiliary textual metadata as a complementary modality. To our knowledge, AuxDet is the first IRSTD framework to introduce and operationalize metadata as a first-class signal for scene-aware target detection.
\end{enumerate}

\subsection{Metadata-Enhanced Visual Recognition}

Auxiliary metadata, such as geographic location, observation time, and viewing angle, has proven valuable in enhancing semantic understanding in image classification tasks. 
These contextual cues provide high-level priors about environmental conditions, lighting, and scene composition, which are often difficult to infer from raw pixel data alone. 
Tang et al.~\cite{tang2015improving} first demonstrated the value of GPS data in improving classification accuracy, leveraging location-specific priors to disambiguate visually similar categories.
Chu et al.~\cite{chu2019geo} integrated temporal metadata and viewing angles via multi-layer perceptrons (MLPs) to disentangle seasonal variations and perspective distortions.
However, these early methods relied on simplistic fusion strategies, limiting their ability to capture complex interactions between visual and auxiliary data.

To overcome these shortcomings, Yang et al.~\cite{yang2022dynamic} proposed a dynamic MLP approach, using instance-specific weighted projections to map similar image features into distinct feature space regions, thus boosting classification precision.
Xu et al.~\cite{xu2023improving} advanced this further with a multi-stage fusion strategy, integrating early-stage input-level fusion with late-stage dynamic MLP structures.
Gao et al.~\cite{gao2025gasc} emphasized early-stage fusion to retain fine-grained geographic details, striking a balance between computational efficiency and contextual richness.
These developments highlight auxiliary information as a source of high-level semantic priors.

It is clear that auxiliary information can explicitly introduce image-related features, which are often crucial high-level semantic information in many tasks. However, in the field of infrared small target detection, there has been no relevant application. 
To the best of our knowledge, our work is the first to introduce and systematically model metadata as an auxiliary signal in the IRSTD domain.

%% file: journal-contents/5-method.tex

\begin{figure*}[hbtp]
\vspace{-1\baselineskip}
  \centering
  \includegraphics[width=0.98\linewidth]{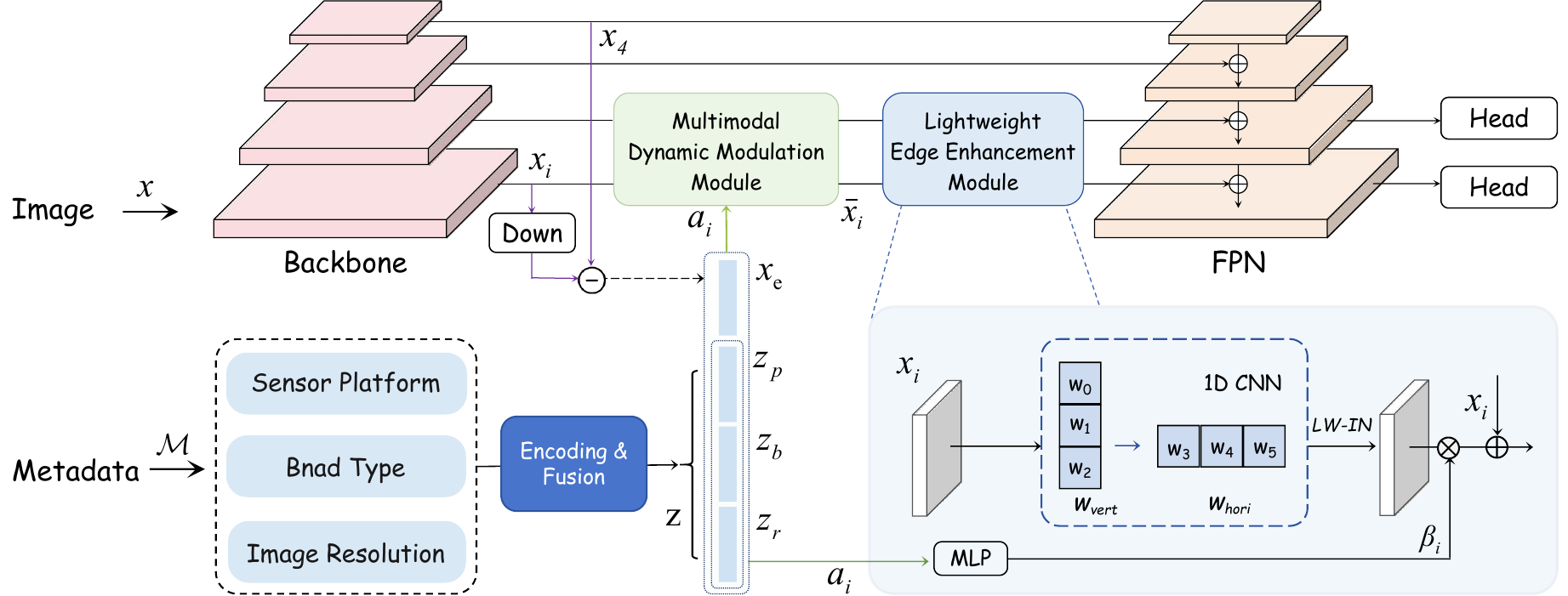}
  \vspace{-0.5\baselineskip}
  \caption{
     Overall architecture of AuxDet.  Infrared images are processed through a backbone network to extract hierarchical features, while auxiliary metadata undergo pre-encoding and fusion via MLPs. The two modalities enter into M2DM module to achieve dynamic feature interaction and instance-level spatial-channel modulation, obtaining modulated features that are subsequently refined by LEEM. The optimized features are fused through the Feature Pyramid Network (FPN) and fed into the detection head for target localization. 
  }
  \label{fig:net}
\vspace{-1\baselineskip}
\end{figure*}

\section{Method} \label{sec:method}

\subsection{Overall Architecture}

As illustrated in Fig.~\ref{fig:net}, the proposed AuxDet framework establishes a novel multimodal architecture for omni-domain IRSTD, building upon the Cascade RPN \citep{vu2019cascade} as the baseline.
Our architecture introduces three dedicated components that synergistically address cross-domain generalization challenges: (1) An auxiliary metadata processing stream that transforms heterogeneous sensor parameters into latent semantic representations, (2) A hierarchical fusion mechanism for vision-metadata co-adaptation, and (3) A edge-aware refinement pipeline for target detail recovery.

The pipeline initiates with parallel feature extraction pathways, a conventional CNN backbone for visual pattern mining and a metadata pre-encoder that embeds physical sensor characteristics into high-dimensional manifolds through learnable tensor projections. These complementary representations undergo alignment in Multi-Modal Dynamic Modulation Module. The modulated features then enter our Lightweight Edge Enhancement Module, which deploys phased 1D convolutional blocks along spatial-channel dimensions to amplify high-frequency target signatures.



\begin{figure}[ht]
  \centering
  \includegraphics[width=0.98\linewidth]{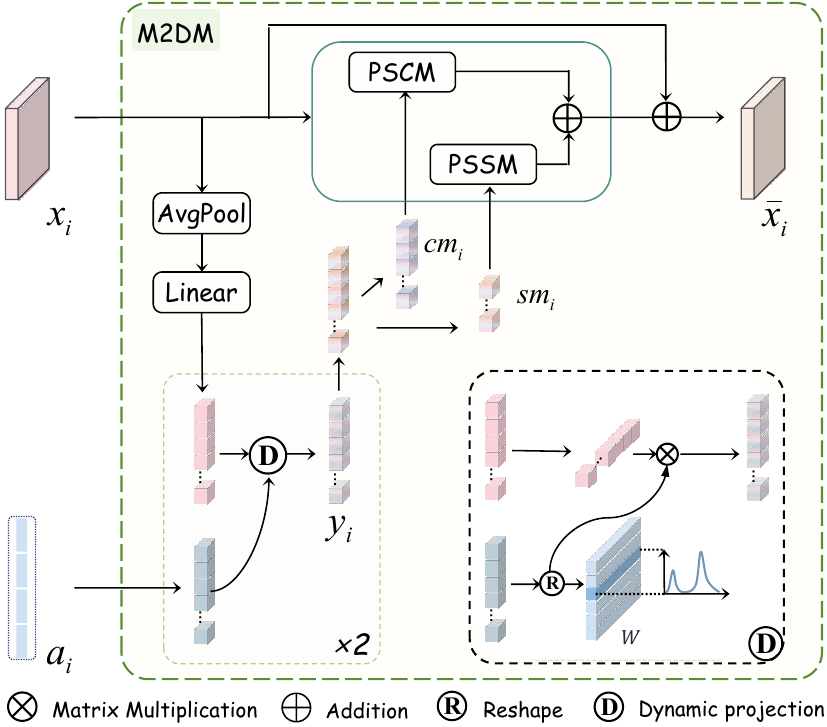}
  \includegraphics[width=0.98\linewidth]{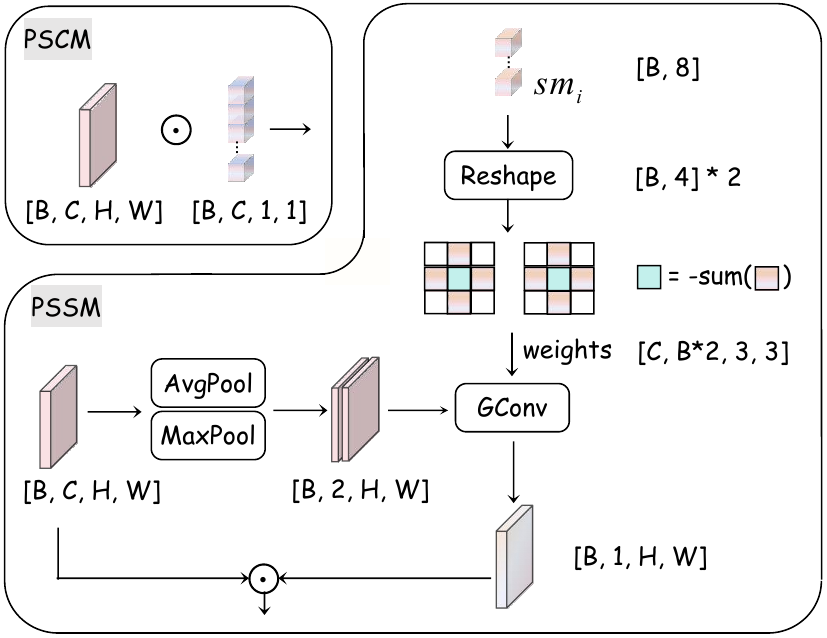}
  \caption{
    Architecture of the M2DM module.
  }
  \label{fig:net01}
\end{figure}

\subsection{Auxiliary Feature Processing and Fusion}

Given an input image $\mathbf{x} \in \mathbb{R}^{3\times H\times W}$, we extract hierarchical visual features $\{\mathbf{x}_i\}_{i=1}^4$ through the backbone, where shallow layers ($i=1,2$) preserve critical spatial details for small target localization, and thus become the embedded layers for our core modules. Our multimodal framework introduces parallel processing of heterogeneous auxiliary metadata $\mathcal{M} = \{platform, band, resolution\}$ via the Multi-Modal Dynamic Modulation module, establishing cross-modal correspondences in the latent space.

\subsubsection{Heterogeneous Metadata Encoding}
To address the multimodal nature of auxiliary information $\mathcal{M}$, we devise a structured encoding strategy. 
Categorical variables undergo one-hot encoding followed by modality-specific projection:
\begin{equation}
\textbf{z}_p = \mathcal{F}_p(\mathrm{OneHot}(platform)),
\end{equation}
\begin{equation}
\textbf{z}_b = \mathcal{F}_b(\mathrm{OneHot}(band)),
\end{equation}
where $\mathcal{F}_p, \mathcal{F}_b$ denote learnable linear transforms. For continuous resolution parameters $(w,h)$, we implement periodic function embedding to capture scale-invariant patterns:
\begin{equation}
\textbf{z}_r = \bigoplus_{k\in\{\sin,\cos\}} \left[ k(\frac{w}{\sigma_w}), k(\frac{h}{\sigma_h}), k(\frac{w}{h}) \right],
\end{equation}
where $\sigma_w, \sigma_h$ represent dataset-specific normalization factors. The multimodal embeddings undergo modality projection operations to align with visual feature space:
\begin{equation}
\textbf{z} = \mathbf{\Psi}_z \left( \bigoplus \left[ \mathcal{F}_r(\textbf{z}_r), \mathcal{F}_p(\textbf{z}_p), \mathcal{F}_b(\textbf{z}_b) \right] \right),
\end{equation}
where $\bigoplus$ denotes channel concatenation and $\mathbf{\Psi}_z$ implements residual fusion through stacked MLP layers with dropout regularization.

\subsubsection{Cross-Modal Feature Compensation}
We believe that due to the characteristics of small targets, detection models tend to focus on shallow information, with insufficient utilization of deep semantic information. Although auxiliary feature information can provide global semantic information to the model and serve as implicit cues for detection tasks, the edge and detail information of small targets still needs further enhancement. In other words, we need to explore the potential of the multimodal path in compensating for the lack of deep-level detail image features. 

Based on this idea, we propose a layer-by-layer differential detail amplifier.
As illustrated in Fig.~\ref{fig:net}, shallow features $\textbf{x}_i$ undergo strided convolution $\mathcal{C}_{\downarrow}$ for spatial alignment, generating pseudo-deep features that preserve high-frequency gradients:
\begin{equation}
\Delta\textbf{x} = \mathcal{C}_{\downarrow}(\textbf{x}_i) - \textbf{x}_4,
\end{equation}
This design explicitly captures geometric deformation information across layers and aggregates high-frequency detail components from shallow networks with information expression from deep networks.

The residual tensor is then non-linearly mapped through a bottleneck MLP $\mathcal{F}_e$ to generate detail-enhanced compensation features:
\begin{equation}
\textbf{x}_e = \mathcal{F}_e(\Delta\textbf{x}), \quad \mathcal{F}_e \in \mathbb{R}^{C\times C/4}.
\end{equation}

Final feature augmentation combines metadata guidance and visual rectification:
\begin{equation}
\mathbf{a}_i = \mathbf{\Psi}_{\text{aux}}\left(\bigoplus\left[\Delta\textbf{x}, \textbf{z}\right]\right),
\end{equation}
where $\mathbf{\Psi}_{aux}$ implements three-stage residual MLP blocks.

This fusion mechanism has dual advantages: the global semantic feature provides contextual priors and decision-level assistance for detection, while the compensation detail feature strengthens the network's response to subpixel-level edge features through cross-layer gradient propagation, effectively alleviating the dilution of detail information in deep features.
\subsection{Multi-Modal Dynamic Modulation Module}

As shown in Fig.~\ref{fig:net01}, our module establishes an instance-adaptive fusion paradigm through conditional feature transformation.
Given aligned visual feature $\mathbf{x}_i$ and auxiliary feature $\mathbf{a}_i$ , we first project them into latent interaction space via average pooling and learnable linear bottleneck $\mathcal{P}_\theta \in \mathbb{R}^{C \times D}$ where $D = \lfloor C/2 \rfloor$, then progressively refine the representations through two-stage dynamic MLP transformations~\cite{yang2022dynamic}. This cascaded architecture enables iterative cross-modal attention:
\begin{equation}
\mathbf{y}'_i = \mathcal{F}_{dyn}^{(2)}\left(\mathcal{F}_{dyn}^{(1)}(\mathcal{P}_\theta(\mathbf{y}_i))\right),
\end{equation}
where each $\mathcal{F}_{dyn}^{(k)}$ implements dynamic parameter prediction through a lightweight network that generates transformation weights from the auxiliary feature. The resultant tensor is decomposed into instance-specific modulation parameters:
\begin{equation}
\begin{array}{c}
[\mathbf{cm}_i, \mathbf{sm}_i] = \mathrm{Split}(\mathbf{y}'_i), \\[1.2ex]
\mathbf{cm}_i \in \mathbb{R}^{B\times C},\quad \mathbf{sm}_i \in \mathbb{R}^{B\times 8},
\end{array}
\end{equation}
Unlike conventional batch-shared modulation, our per-sample parameterization preserves batch dimension $B$ to enable instance-wise domain adaptation based on metadata.


\noindent\textbf{Per-Sample Channel Modulation (PSCM):}
The channel modulation applies element-wise affine transformations conditioned on imaging metadata:
\begin{equation}
\mathbf{x}_i^c = \mathbf{x}_i \otimes \sigma(\mathbf{cm}_i), 
\end{equation} 
where $\otimes$ denotes Hadamard product and $\sigma(\cdot)$ represents the sigmoid activation function.

\noindent\textbf{Per-Sample Spatial Modulation (PSSM):} 
We reformulate spatial modulation parameters into edge-aware operators through zero-corner constrained kernel construction:
\begin{equation}
\textbf{K}^{(1)}_i = 
\begin{bmatrix}
    0 & \textbf{\textit{sm}}_{i}^1 & 0 \\
    \textbf{\textit{sm}}_{i}^2 & -\sum_{k=1}^4 \textbf{\textit{sm}}_{i}^k & \textbf{\textit{sm}}_{i}^3 \\
    0 & \textbf{\textit{sm}}_{i}^4 & 0
\end{bmatrix},
\end{equation}
\begin{equation}
\textbf{K}^{(2)}_i = 
\begin{bmatrix}
    0 & \textbf{\textit{sm}}_{i}^5 & 0 \\
    \textbf{\textit{sm}}_{i}^6 & -\sum_{k=5}^8 \textbf{\textit{sm}}_{i}^k & \textbf{\textit{sm}}_{i}^7 \\
    0 & \textbf{\textit{sm}}_{i}^8 & 0
\end{bmatrix},
\end{equation}
This constrained design emulates Laplacian-of-Gaussian filtering for high-frequency enhancement. 
For hardware-efficient computation of sample-wise spatial modulation, we decouple batch and channel dimensions through grouped convolution, enabling sample-wise parallel processing.

To enable lightweight modulation, we employ dual-stream max pooling $\mathcal{P}_{max}(\cdot)$ and average pooling $\mathcal{P}_{avg}(\cdot)$ for multi-scale context aggregation:
\begin{equation}
\mathbf{f}_{sm_i} = \sum_{k=1}^2 \mathcal{G}_k \ast \mathbf{f}_{pool_i}^{(k)}, \quad \mathcal{G}_k \in \mathbb{R}^{G\times 1\times 3\times 3},
\end{equation}
where $\mathbf{f}_{pool_i} = \mathcal{P}_{avg}(\mathbf{x}_i) \oplus \mathcal{P}_{max}(\mathbf{x}_i)$ captures multi-scale context through parallel pooling streams. 
$\mathcal{G}_k$ denotes group convolution with $k$ subgroups, and $\ast$ is the convolution operation.
The modulated features undergo gated residual fusion:
\begin{equation}
\mathbf{x}_i^s = \mathbf{x}_i + \sigma(\mathbf{f}_{sm_i}) \otimes \mathbf{x}_i.
\end{equation}

The final representation integrates modulated features through residual pathways:
\begin{equation}
\overline{\mathbf{x}}_i = \mathbf{x}_i + \mathbf{x}_i^c + \mathbf{x}_i^s.
\end{equation}

\subsection{Lightweight Edge Enhancement Module}

A pivotal challenge in IRSTD lies in the vulnerability of small target features to being overwhelmed by intricate background clutter. To counter this, we introduce a Lightweight Edge Enhancement Module aimed at refining fused features and restoring subtle target details, while minimizing the introduction of additional parameters. 

To achieve this, we propose a decomposed edge-aware unit that leverages a vertical-horizontal 1D convolution chain to extract edge responses. By decoupling vertical and horizontal edge information, this design strikes an effective balance between computational efficiency and precision in edge delineation. For an input feature map \(\textbf{x} \in \mathbb{R}^{C \times H \times W}\), the edge enhancement process is defined as:
\begin{equation}
\begin{aligned}
\textbf{x}_{\text{vert}} &= \sigma(\alpha) \odot \text{ReLU}({W}_{3\times1} \ast \textbf{x}), \\
\textbf{x}_{\text{hori}} &= {W}_{1\times3} \ast \textbf{x}_{\text{vert}},
\end{aligned}
\label{eq:edge_conv}
\end{equation}
where \({W}_{3\times1}\) and \({W}_{1\times3}\) denote learnable convolutional kernels, \(\alpha \in \mathbb{R}\) acts as a gating parameter.
Inspired by Marr-Hildreth edge detection theory, we initialize kernels with discrete Laplacian values to establish stable geometric priors:
\begin{equation}
\begin{cases}
{W}_{3\times1}^{(0)} = \frac{1}{4} \cdot [1,-2,1]^{T} \odot {I}_C \\ 
{W}_{1\times3}^{(0)} = \frac{1}{4} \cdot [1,-2,1] \odot {I}_C
\end{cases},
\label{eq:weight_init}
\end{equation}
where ${I}_C \in \mathbb{R}^{C \times C}$ represents the identity matrix that preserves channel-wise independence during convolution kernel initialization and the 1D Laplacian vector is repeated across all $C$ channels. 

This initialization strategy simulates the edge sensitivity of local difference operators. 
When the vertical kernel \( {W}_{3 \times 1} \) and horizontal kernel \( {W}_{1 \times 3} \) are cascaded for convolution, the equivalent 3×3 convolution kernel can be represented as the 2D Laplacian-of-Gaussian (LoG) filtering:
\begin{equation}
{W}_{3\times1} \ast {W}_{1\times3} = \frac{1}{16} \begin{bmatrix}
1 & -2 & 1 \\
-2 & 4 & -2 \\
1 & -2 & 1
\end{bmatrix} \odot {I},
\label{eq:combined_kernel}
\end{equation}
The kernel satisfies the second-order differential properties of the discrete Laplacian operator: the center element value (4/16 = 0.25) equals the negative sum of the values of the surrounding eight neighborhood elements (\( (-2) \times 4 + 1 \times 4 = -4 \Rightarrow -4/16 = -0.25 \)). This mathematical property provides the following advantages:

\begin{enumerate}
\item \textbf{Curvature sensitivity}: When the gray difference between the central pixel and the surrounding neighborhood changes nonlinearly (e.g., in edge transition areas), the operator generates a large response.
This formulation ensures intrinsic curvature sensitivity, yielding robust edge responses from the initial training stages. Consequently, our model achieves superior small target awareness, outperforming most baselines (AP\textsubscript{50}: 73.0\%+) within just 6 epochs.
\item \textbf{Zero-sum property}: In uniform regions, it satisfies \( \sum_{i,j} W_{i,j} = 0 \), suppressing artifacts caused by lighting changes.
\item \textbf{Isotropy}: The rotationally symmetric kernel structure is insensitive to edge direction, ensuring balanced detection of edges at multiple angles.
\end{enumerate}

Through L2-norm scaling (the \( \frac{1}{16} \) factor), the initial gradient norm is constrained, effectively avoiding gradient explosion at the beginning of training.


Then, we apply \textbf{Lightweight Instance Normalization (LW-IN)} per sample to mitigate batch normalization’s domain shift issues, ensuring robust performance across diverse imaging scenarios. The output is expressed as: 
\begin{equation}
\overline{\mathbf{x}} = \mathbf{x} + \gamma \odot \left(\frac{\textbf{x}_{\text{hori}} - \mu_{\text{space}}}{\sqrt{\sigma_{\text{space}}^2 + \epsilon}}\right),
\label{eq:lwin}
\end{equation}
where $\mu_{\text{space}}$ and $\sigma_{\text{space}}^2$ denote the spatial mean and variance computed across each channel, $\gamma \in \mathbb{R}^C$ is the learnable scale parameter, and $\epsilon=10^{-5}$ is used for numerical stability. This design preserves feature distribution stability across heterogeneous imaging conditions 
by eliminating cross-sample statistical coupling.




%% file: journal-contents/6-experiment.tex

\section{Experiments} \label{sec:experiment}

\subsection{Experimental Settings} \label{subsec:setting}
\textit{1) {Datasets}: }To thoroughly evaluate the effectiveness of the proposed AuxDet framework, we adopt the publicly available WideIRSTD-Full Dataset~\cite{Li2024First}.
It integrates seven prominent public datasets, namely SIRST-V2~\cite{dai2023one}, IRSTD-1K~\cite{zhang2022isnet}, IRDST~\cite{sun2023receptive}, NUDT-SIRST~\cite{li2022dense}, NUDT-SIRST-Sea~\cite{wu2023mtu}, NUDT-MIRSDT~\cite{li2025direction}, and Anti-UAV~\cite{jiang2023anti}, augmented by an additional dataset from the National University of Defense Technology.
It spans diverse imaging platforms, spectral bands, and target types, encompassing both simulated and real-world data with significant variability, which includes 9,000 training images with corresponding metadata. And we further split 7,000 images for training and 2,000 images for validation. 

\textit{2) {Evaluation Metrics}: }We employ Recall and Average Precision at IoU threshold 0.5 ($AP_{50}$) as primary metrics. 

\textit{3) {Implementation Details}:}
The proposed method is implemented using PyTorch 2.0.1 and MMDetection 3.3.0, with an NVIDIA RTX 3090 acceleration via CUDA 11.7. The maximum training duration is 12 epochs, with the learning rate initialized at 0.005 using a linear warmup for the first 500 iterations.
Given the prevalence of large-scale images in the dataset, the batch size is set as 4 and all images are resized to $1024 \times 1024$. 

For loss functions:
Smooth L1 loss is adopted for bbox regression in the ROI Head.
IoU loss serves as the refinement loss for two-stage candidate box regression.
CrossEntropy loss handles classification in the second stage.

\begin{table*}[h]
  \renewcommand\arraystretch{1.3}
  \small
  \centering
  \caption{Comparison with Other State-of-the-art methods on WideIRSTD-Full Dataset.}
  \label{tab:WideIRSTD_comparison}
  \setlength{\tabcolsep}{5.5pt}
  \begin{tabular}{l|c|c|cccc}  
    \toprule
    \multicolumn{1}{c|}{\multirow{2}{*}{\textbf{Method}}}
    & \multirow{2}{*}{\textbf{Venue}} & \multirow{2}{*}{\textbf{Backbone}} & \multicolumn{4}{c}{\textbf{WideIRSTD-Full Dataset}} \\
    & & & FLOPs $\downarrow$ & Params $\downarrow$ & AP\textsubscript{50} (\%)  $\uparrow$ & Recall (\%) $\uparrow$ \\
    \midrule
    \multicolumn{7}{l}{\textit{General Object Detection} \quad 
    
    $\triangleright$~\textit{One-stage} \quad $\triangleright$~\textit{Two-stage} \quad $\triangleright$~\textit{End-to-End}} \\
    
    \midrule
    RetinaNet~\cite{lin2017focal} & ICCV$^{\textbf{17}}$ & ResNet50 & 0.197T & 36.330M & 60.7 & 74.8 \\ 
    FCOS~\cite{tian2019fcos} & ICCV$^{\textbf{19}}$ & ResNet50 & 0.194T & 32.113M & 49.2 & 58.3 \\
    CenterNet~\cite{zhou2019objects} & arXiv$^{\textbf{19}}$ & ResNet50 & 0.194T & 32.111M & 10.9 & 24.5 \\
    GFL~\cite{li2020generalized} & NeurIPS$^{\textbf{20}}$ & ResNet50 & 0.197T & 32.258M & 48.5 & 60.2 \\
    ATSS~\cite{zhang2020bridging} & CVPR$^{\textbf{20}}$ & ResNet50 & 0.194T & 32.113M & 50.1 & 60.6 \\
    AutoAssign~\cite{zhu2020autoassign} & arXiv$^{\textbf{20}}$ & ResNet50 & 0.195T & 36.244M & 51.2 & 62.1 \\
    YOLOF~\cite{chen2021you} & CVPR$^{\textbf{21}}$ & ResNet50 & 71.227G & 44.160M & 17.5 & 21.9 \\
    DyHead~\cite{dai2021dynamic} & CVPR$^{\textbf{21}}$ & ResNet50 & 0.105T & 38.890M & 0.6 & 15.9 \\
    TOOD~\cite{feng2021tood} & ICCV$^{\textbf{21}}$ & ResNet50 & 0.191T & 32.018M & 53.2 & 61.9 \\
    DDOD~\cite{chen2021disentangle} & ACM MM$^{\textbf{21}}$ & ResNet50 & 0.172T & 32.196M & 52.2 & 62.0 \\
    \midrule
    Faster R-CNN~\cite{ren2015faster} & NeurIPS$^{\textbf{15}}$ & ResNet50 & 0.200T & 41.348M & 31.2 & 33.5 \\
    Cascade R-CNN~\cite{cai2019cascade} & TPAMI$^{\textbf{19}}$ & ResNet50 & 0.228T & 69.152M & 35.8 & 37.5 \\
    Grid R-CNN~\cite{lu2019grid} & CVPR$^{\textbf{19}}$ & ResNet50 & 0.313T & 64.467M & 32.2 & 34.8 \\
    Cascade RPN~\cite{vu2019cascade} & NeurIPS$^{\textbf{19}}$ & ResNet50 & 0.190T & 42.527M & 74.4 & 84.6 \\
    Libra R-CNN~\cite{pang2019libra} & CVPR$^{\textbf{19}}$ & ResNet50 & 0.209T & 41.611M & 30.7 & 33.8 \\
    Dynamic R-CNN~\cite{zhang2020dynamic} & ECCV$^{\textbf{20}}$ & ResNet50 & 0.200T & 41.348M & 35.0 & 37.2 \\
    SABL~\cite{wang2020side} & ECCV$^{\textbf{20}}$ & ResNet50 & 0.198T & 36.357M & 64.1 & 77.8 \\
    \midrule
    Deformable DETR~\cite{zhu2021deformable} & ICLR$^{\textbf{21}}$ & ResNet50 & 0.189T & 40.099M & 33.1 & 49.8 \\
    Conditional DETR~\cite{meng2021conditional} & ICCV$^{\textbf{21}}$ & ResNet50 & 98.099G & 43.448M & 0.0 & 0.2 \\
    Sparse R-CNN~\cite{sun2021sparse} & CVPR$^{\textbf{21}}$ & ResNet50 & 0.146T & 0.106G & 40.4 & 62.1 \\
    DINO~\cite{zhang2022dino} & ICLR$^{\textbf{23}}$ & ResNet50 & 0.268T & 47.540M & 67.1 & 82.8 \\
    DAB-DETR~\cite{liu2022dabdetr} & ICLR$^{\textbf{22}}$ & ResNet50 & 99.511G & 43.702M & 17.6 & 31.9 \\

    \midrule
    
    \multicolumn{6}{l}{\textit{Small Target Specialized}}  \\
    \midrule
    ALCNet~\cite{dai2021attentional} & TGRS$^{\textbf{21}}$ & ResNet20 & 6.047G & 42.700M & 36.3 & 69.7 \\
    NWD~\cite{xu2022detecting} & ISPRS$^{\textbf{22}}$ & ResNet50 & 0.235T & 0.123G & 48.5 & 53.6 \\
    RFLA~\cite{xu2022rfla} & ECCV$^{\textbf{22}}$ & ResNet50 & 0.235T & 0.123G & 71.8 & 80.6 \\
    OSCAR~\cite{dai2023one} & TGRS$^{\textbf{23}}$ & ResNet50 & 0.327T & 30.730M & 40.2 & 65.4 \\
    DNANet~\cite{li2022dense} & TIP$^{\textbf{23}}$ & - & 0.228T & 4.697M & 73.9 & 46.9 \\
    RDIAN~\cite{TGRS23RDIAN} & TGRS$^{\textbf{23}}$ & -  & 0.059T & 0.217M & 62.5 & 51.8 \\
    DQ-DETR~\cite{huang2024dq} & ECCV$^{\textbf{24}}$ & ResNet50 & 1.476T & 58.680M & 70.1 & 81.3 \\
    EFLNet~\cite{yang2024eflnet} & TGRS$^{\textbf{24}}$ & - & 0.102T & 38.336M & 73.7 & 68.1 \\
    PConv~\cite{yang2025pinwheel} & AAAI$^{\textbf{25}}$ & - & 0.012T & 2.922M & 71.7 & 64.5 \\
    \midrule
    \rowcolor[rgb]{0.9,0.9,0.9} \textbf{Ours} & - & ResNet50 & 0.215T & 45.279M & \textbf{77.9} & \textbf{87.2} \\
    \bottomrule
  \end{tabular}
\end{table*}

\subsection{Comparison with State-of-the-Arts} 
We conduct rigorous comparisons with 31 representative detectors, including general object detectors (one-stage, two-stage, and end-to-end) and small-target specialized detectors.
As shown in Tab.~\ref{tab:WideIRSTD_comparison}, our analysis reveals three principal observations:

\textit{1) Generic Detector Inadequacy}: 
Despite architectural advances, generic detectors for natural images exhibit catastrophic failures (avg. AP\textsubscript{50}$<$45\%) due to inherent scale biases and the inability to effectively capture small targets. 
Among these methods, Cascade RPN emerges as the best performer (74.4\% AP\textsubscript{50}) through its multi-stage anchor refinement and adaptive convolution that mitigates feature misalignment. DINO's contrastive denoising (82.8\% Recall) improves small object sensitivity but suffers from transformer's intrinsic quadratic memory complexity.

\textit{2) Limitations of Single-Modal Specialized Models}:
Domain-specific approaches reveal critical tradeoffs. For example, EFLNet's adaptive threshold focal loss (73.7\% AP\textsubscript{50}) successfully suppresses background but over-prunes channel dimensions, collapsing on faint targets (68.1\% Recall).

\textit{3) AuxDet's Multi-Modal Breakthrough:}
Our framework achieves SOTA 77.9\% AP\textsubscript{50} and 87.2\% Recall, while still maintaining superior parameter efficiency compared to EFLNet.
This advancement hinges on metadata-guided modulation, which dynamically refines feature representations for robust detection across omni-domain IRSTD.

\subsection{Visual Analysis} \label{subsec:visualization}

\begin{figure*}[hbtp]
\vspace{-1\baselineskip}
  \centering
  \includegraphics[width=0.98\linewidth]{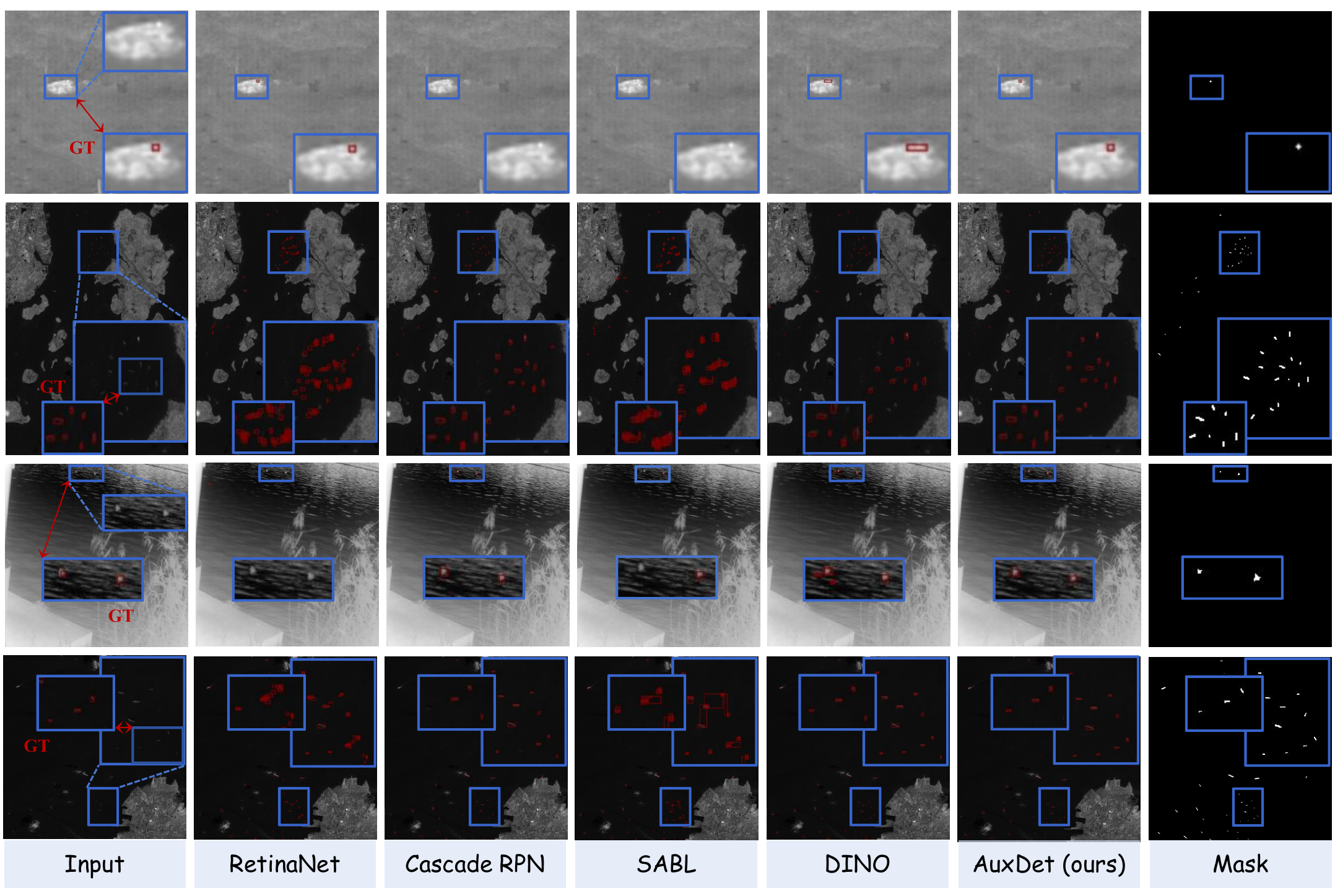}
  \vspace{-0.5\baselineskip}
  \caption{
    Visualization of detection results comparing AuxDet with RetinaNet, Cascade RPN, SABL, DINO.
  }
  \label{fig:vs}
\vspace{-1\baselineskip}
\end{figure*}

As illustrated in Fig.~\ref{fig:vs}, we compare our AuxDet with the top-4 performing methods. In scenarios where targets are submerged in background clutter (first row), Cascade RPN and SABL exhibit missed detections, while DINO demonstrates significant bounding box regression errors. For dense low-contrast targets (second and fourth rows), RetinaNet and SABL completely fail to localize small targets, whereas Cascade RPN and DINO suffer from both missed detections and false alarms. In wave-texture interference scenarios (third row), SABL misses partial targets, RetinaNet shows severe false positives, and DINO misidentifies waves as targets. In contrast, our method achieves precise target detection with accurate bounding box regression, delivering visually superior results.

\begin{figure*}[hbtp]
\vspace{-1\baselineskip}
  \centering
  \includegraphics[width=0.98\linewidth]{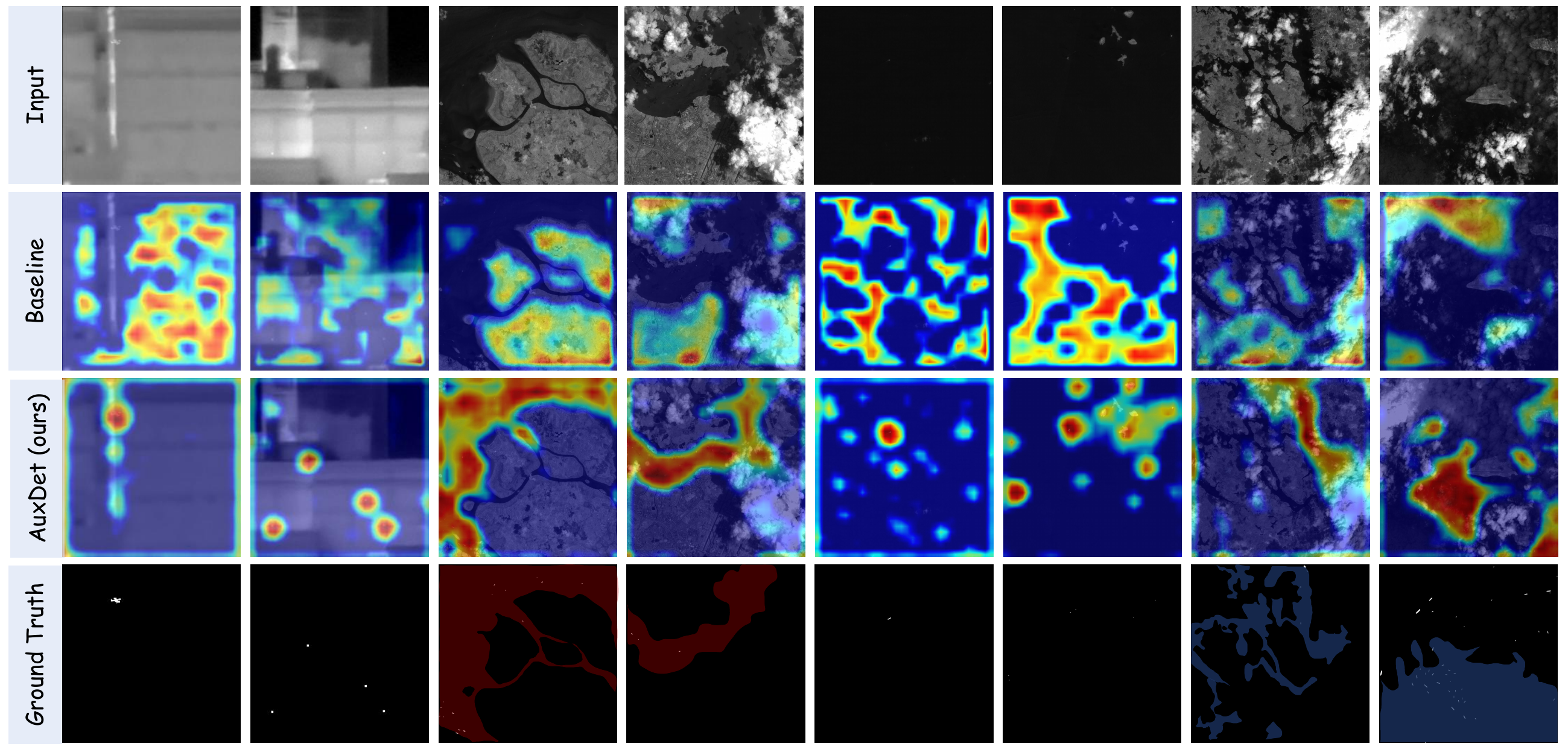}
  \vspace{-0.5\baselineskip}
  \caption{
    Feature visualization comparison. White masks annotate small targets, where comparative analysis demonstrates our method's enhanced small-target focusing capability over the baseline. In ground truth annotations, we use red to mark distinct contiguous maritime regions and blue to denote barely discernible maritime areas under heavy cloud interference. Visually, our method achieves semantic-level localization of small targets in complex backgrounds, precisely delineating their spatial extent with auxiliary information guidance.
  }
  \label{fig:visualization}
\vspace{-1\baselineskip}
\end{figure*}

To further validate the global semantic-level guidance enabled by our auxiliary metadata-driven modulation, we visualize the output features of the neck, as shown in Fig.~\ref{fig:visualization}. In relatively simple land-based small target detection scenarios (columns 1–2), our method effectively focuses on the small targets. In space-based imagery, which often involves complex scenes encompassing atmosphere, land, and ocean, our approach is still capable of achieving semantic-level localization of small targets against complex backgrounds.

In the visualization, we use red and blue masks to indicate clearly distinguishable sea areas and visually ambiguous regions with strong cloud interference in maritime detection scenes, respectively. As observed, when the sea and land areas are relatively balanced and the sea is clearly visible to the naked eye (columns 3–4), our method provides a strong prior localization toward the sea regions. When the sea area is minimal (columns 5–6) and small land masses are visually similar to the targets, our method is still able to directly focus on the actual targets, effectively reducing false alarms. In highly entangled scenarios where sea, atmosphere, and land are difficult to distinguish (columns 7–8), our method not only maintains strong prior localization toward the sea, but also successfully highlights target regions partially obscured by clouds and barely discernible to the human eye.

Overall, our approach demonstrates superior target localization performance compared to the baseline, underscoring the role of auxiliary metadata in delivering high-level semantic guidance and facilitating robust adaptability across diverse domains in Omni-IRSTD.



\begin{table*}[htbp]
  \setlength{\abovecaptionskip}{0cm}
  \renewcommand\arraystretch{1.3}
  \small
  \centering
  \caption{Evaluation of general detectors and small-target specialized detectors with and without AuxDet.}
  \label{tab:plug_comparison_revised}
  \setlength{\tabcolsep}{6pt}
    \begin{tabular}{l|cc|cc}
      \toprule
      \multicolumn{1}{c|}{\multirow{2}{*}{\textbf{Method}}} & \multicolumn{2}{c|}{\textbf{Before (Baseline)}} & \multicolumn{2}{c}{\textbf{After (w/ AuxDet)}} \\
       & AP\textsubscript{50} (\%) $\uparrow$ & Recall (\%) $\uparrow$ & AP\textsubscript{50} (\%) $\uparrow$ & Recall (\%) $\uparrow$ \\
      \midrule
      \multicolumn{5}{l}{\textit{General Object Detection}} \\
      \midrule
      Faster R-CNN~\cite{ren2015faster}     & 31.2 & 33.5 & 34.8 (+3.6) & 36.6 (+3.1) \\
      Cascade R-CNN~\cite{cai2019cascade}   & 35.8 & 37.5 & 37.5 (+1.7) & 39.0 (+1.5) \\
      Dynamic R-CNN~\cite{zhang2020dynamic} & 35.0 & 37.2 & 36.8 (+1.8) & 38.8 (+2.0) \\
      Grid R-CNN~\cite{lu2019grid}          & 32.2 & 34.8 & 33.8 (+1.6) & 36.0 (+1.2) \\
      Libra R-CNN~\cite{pang2019libra}      & 30.7 & 33.8 & 32.2 (+1.5) & 34.8 (+1.0) \\
      SABL~\cite{wang2020side}              & 64.1 & 77.8 & 65.9 (+1.8) & 79.3 (+1.5) \\
      \midrule
      \multicolumn{5}{l}{\textit{Small Target Specialized}} \\
      \midrule
      OSCAR~\cite{dai2023one}               & 40.2 & 65.4 & 47.5 (+7.3)  & 71.5 (+6.1) \\
      RFLA~\cite{xu2022rfla}                & 71.8 & 80.6 & 73.5 (+1.7)  & 84.1 (+3.5) \\
      NWD~\cite{xu2022detecting}            & 48.5 & 53.6 & 61.1 (+12.6) & 69.2 (+15.6) \\
      ALCNet~\cite{dai2021attentional}      & 36.3 & 69.7 & 42.2 (+5.9)  & 74.0 (+4.3) \\
      \bottomrule
    \end{tabular}
\end{table*}

\subsection{Analysis of the Integration Flexibility of the Proposed Framework}
To demonstrate the generalizability of our approach, we integrate the proposed AuxDet framework as a general-purpose auxiliary module into several representative detection architectures. Specifically, we consider both generic object detectors (e.g., Faster R-CNN, Cascade R-CNN) and small target–oriented methods (e.g., OSCAR, NWD).

As summarized in Tab.~\ref{tab:plug_comparison_revised}, AuxDet consistently enhances detection performance across all models, yielding improvements in both AP\textsubscript{50} and Recall. These gains are observed regardless of the underlying architectural design or training strategies, reflecting the framework’s broad applicability.

Notably, the performance improvements on small target–specialized detectors further indicate that even methods tailored to small-object characteristics can benefit from metadata-guided adaptation. This confirms that AuxDet introduces complementary capabilities that strengthen domain awareness and target discrimination across diverse sensing conditions.

\subsection{Comparison with Transformer-based Methods}

While our primary comparisons in this paper are conducted against representative CNN-based approaches, Transformer-based models have recently demonstrated promising potential in the field of infrared small target detection (IRSTD), owing to their strong global modeling capabilities and adaptability to complex visual contexts.

To further validate the effectiveness and generalizability of our proposed framework, we conduct two complementary experiments:
(1) we integrate AuxDet into Transformer-based backbones to evaluate its compatibility and performance gains in modern vision architectures;
(2) we directly compare our method with several state-of-the-art Transformer-based IRSTD methods.

\subsubsection{Integration with Transformer Backbones}
We replace the CNN backbone with two popular Transformer architectures—PVTv2~\cite{wang2022pvt} and Swin-T~\cite{liu2021swin}—and evaluate the detection performance before and after applying AuxDet. As shown in Tab.~\ref{tab:transformer_backbone}, AuxDet continues to bring consistent improvements across both backbones, achieving +2.2 AP\textsubscript{50} and +3.1 Recall on PVTv2-B3, and +2.1 AP\textsubscript{50} and +3.2 Recall on Swin-T. These results further demonstrate the general-purpose nature of AuxDet, which can effectively complement Transformer-based visual representations.

\begin{table}[htbp]
  \setlength{\abovecaptionskip}{0cm}
  \renewcommand\arraystretch{1.3}
  \small
  \centering
  \caption{Evaluation of AuxDet with transformer backbones.}
  \label{tab:transformer_backbone}
  \setlength{\tabcolsep}{6pt}
    \begin{tabular}{l|cc|cc}
    \toprule
      \multirow{2}{*}{\textbf{Backbone}} & \multicolumn{2}{c|}{\textbf{Baseline}} & \multicolumn{2}{c}{\textbf{w/ AuxDet}} \\
       & AP\textsubscript{50}(\%)$\uparrow$ & Recall(\%)$\uparrow$ & AP\textsubscript{50}(\%) & Recall(\%) \\
      \midrule
      PVTv2-B3 & 64.7 & 78.0 & \textbf{66.9} & \textbf{81.1} \\
      Swin-T   & 65.4 & 79.2 & \textbf{67.5} & \textbf{82.4} \\
    \midrule
    \end{tabular}
\end{table}

\subsubsection{Comparison with Transformer-based IRSTD Methods}
We also compare AuxDet with several recent Transformer-based IRSTD models, including SCTransNet, DQ-DETR, and DATransNet. As reported in Tab.~\ref{tab:transformer_comparison}, our method achieves superior performance in both AP\textsubscript{50} and Recall, surpassing the best competing method (SCTransNet) by +2.4 AP\textsubscript{50} and +20.8 Recall. This highlights the effectiveness of our metadata-guided scene-aware design in capturing both spatial and contextual cues essential for small target detection, especially in complex omni-domain scenarios.

\begin{table}[htbp]
  \setlength{\abovecaptionskip}{0cm}
  \renewcommand\arraystretch{1.3}
  \small
  \centering
  \caption{Comparison with transformer-based methods.}
  \label{tab:transformer_comparison}
  \setlength{\tabcolsep}{6pt}
    \begin{tabular}{l|cc}
      \toprule
      \textbf{Method} & \textbf{AP\textsubscript{50} (\%)}$\uparrow$ & \textbf{Recall (\%)}$\uparrow$ \\
      \midrule
      SCTransNet~\cite{yuan2024sctransnet} & 75.5 & 66.4 \\
      DQ-DETR~\cite{huang2024dq}           & 70.1 & 81.3 \\
      DATransNet~\cite{hu2025datransnet}   & 56.7 & 43.8 \\
      \rowcolor[rgb]{0.9,0.9,0.9}
      \textbf{Ours}               & \textbf{77.9} & \textbf{87.2} \\
      \bottomrule
    \end{tabular}
\end{table}

In summary, the results from both integration and comparative studies confirm that AuxDet is not only compatible with Transformer-based architectures but also capable of outperforming specialized Transformer models designed for IRSTD. This underscores the robustness and generalization ability of our framework across different model families and detection paradigms.

\subsection{Ablation Study} \label{subsec:ablation}
\subsubsection{{Efficacy of Core Modules}}
To probe the interplay between M2DM and LEEM, we assessed their standalone and joint configurations. As evidenced in Tab.~\ref{tab:ModuleAblation}, their combined deployment achieves a striking +3.2\% AP\textsubscript{50} uplift, validating the architectural coherence and functional potency.
This synergy emerges from M2DM's global scene-aware modulation complementing LEEM's localized edge refinement, collectively enhancing both contextual and structural target cues.

\begin{table}[h]
    \centering
    \caption{Ablation study on core modules M2DM and LEEM. The best results are highlighted in bold.}
    \label{tab:ModuleAblation}
    \renewcommand{\arraystretch}{1.3} 
    \setlength{\tabcolsep}{5pt} 
    \footnotesize 
    \begin{tabular}{cc|cc|c}
        \toprule
        \multicolumn{2}{c|}{\textbf{Module}} & \multicolumn{2}{c|}{\textbf{Performance}} & \multirow{2}{*}{\textbf{Params. (M)}} \\
        M2DM & LEEM & AP\textsubscript{50} (\%) $\uparrow$ & Recall (\%) $\uparrow$ & \\
        \midrule
           &   & 74.7 & 84.5 & 42.527 \\
        \ding{51} &   & 77.2 & 86.7 & 44.367 \\
           & \ding{51} & 77.2 & 86.7 & 42.723 \\
        \rowcolor[rgb]{0.9,0.9,0.9}
        \ding{51} & \ding{51} & \textbf{77.9} & \textbf{87.2} & 45.279 \\
        \bottomrule
    \end{tabular}
\end{table}

    

\subsubsection{{Analysis of Multi-Modal Dynamic Modulation Module}}

We systematically evaluate the effectiveness of dynamic modulation mechanisms by comparing different integration strategies of PSSM and PSCM. 
The ablation study in Table~\ref{tab:PSSM_PSCM} reveals three critical insights:




\begin{itemize}
    \item {Independent Enhancement Pathways}: PSSM's edge-sensitive spatial warping (+1.1\% AP\textsubscript{50}) compensates for target boundary erosion in thermal diffusion scenarios. PSCM's cross-channel contextual modeling increases the performance by 1.0\% (74.7\% $\rightarrow$ 75.7\%).
    
    \item {Parallel Fusion Superiority}: Parallel fusion achieves 77.2\% AP\textsubscript{50} via complementary interaction, namely spatial modulation preserves structural integrity while channel attention suppresses clutter harmonics.
    
    \item {Parameter-Efficient Design}: Shared dynamic projection layers reduce duplicate parameters, achieving 1.84M overhead (only 4.1\% of baseline).
\end{itemize}

This evidence confirms our hypothesis: PSSM's edge-aware spatial modulation and PSCM's context-aware channel attention form orthogonal feature enhancement pathways. Their parallel implementation maximizes mutual reinforcement while avoiding parametric explosion through weight-sharing in dynamic projection layers.
\begin{table}[htbp]
    \centering
    \caption{Ablation study on the M2DM module. The best results are highlighted in bold.}
    \label{tab:PSSM_PSCM}
    \renewcommand\arraystretch{1.3} 
    \setlength{\tabcolsep}{3pt} 
    \footnotesize 
    \begin{tabular}{l|cc|r}
        \toprule
        \multicolumn{1}{c|}{\multirow{2}{*}{\textbf{Module Config.}}} & \multicolumn{2}{c|}{\textbf{Performance}} & \multirow{2}{*}{\textbf{Params. (M)}} \\
         & AP\textsubscript{50} (\%) $\uparrow$ & Recall (\%) $\uparrow$ & \\
        \midrule
        Baseline & 74.7 & 84.5 & 42.527 \\
        PSSM Only & 75.8 & 85.7 & 44.234 \\
        PSCM Only & 75.7 & 85.3 & 44.230 \\
        PSSM $\rightarrow$ PSCM & 76.3 & 86.1 & 44.367 \\
        PSCM $\rightarrow$ PSSM & 76.8 & 86.3 & 44.367 \\
        \rowcolor[rgb]{0.9,0.9,0.9}
        PSSM + PSCM & \textbf{77.2} & \textbf{86.7} & 44.367 \\
        \bottomrule
    \end{tabular}
\end{table}


\subsubsection{Analysis of Lightweight Edge Enhancement}
To substantiate the efficacy and efficiency of the LEEM, we benchmarked it against a $3\times3$ edge convolution operator, initialized to mimic LEEM's behavior (per Eq.~(\ref{eq:combined_kernel})). As reported in Tab.~\ref{tab:EdgeConv}, our decomposed 1D convolutional design delivers commensurate performance gains with a parameter increase of less than 1\%.

\begin{table}[htbp]
  \centering
  \caption{Ablation study on LEEM. The best results are highlighted in bold.}
  \label{tab:EdgeConv}
  \renewcommand{\arraystretch}{1.3} 
  \setlength{\tabcolsep}{5pt} 
  \footnotesize 
    \begin{tabular}{l|cc|c}
      \toprule
      \multicolumn{1}{c|}{\multirow{2}{*}{\textbf{Module Config.}}} & \multicolumn{2}{c|}{\textbf{Performance}} & \multirow{2}{*}{\textbf{Params. (M)}} \\
       & AP\textsubscript{50} (\%) $\uparrow$ & Recall (\%) $\uparrow$ &  \\
      \midrule
      Baseline & 74.7 & 84.5 & 42.527 \\
      w/ W$_{3\times3}$ (Standard) & \textbf{77.3} & 86.7 & 43.116 \\
      \rowcolor[rgb]{0.9,0.9,0.9} 
      w/ W$_{1\times3} \rightarrow$ W$_{3\times1}$ & 77.2 & \textbf{86.8} & 42.723 \\
      \bottomrule
    \end{tabular}
\end{table}



\subsubsection{Impact of the Embedding Depth of Core Modules}
In our model design, both the M2DM and the Lightweight Edge Enhancement Module (LEEM) are embedded after the shallow backbone features, which are rich in detailed information.  This placement is not only intuitively reasonable, but also empirically validated through comprehensive experiments.  As shown in Tab.~\ref{tab:EmbedDepth}, we vary the embedding depth of M2DM and LEEM while keeping other parameters and modules unchanged.  The depth is counted from shallow to deep;  for example, a depth of 2 indicates that the core modules are inserted after stage 1 and stage 2 of the backbone, respectively.

\begin{table}[htbp]
  \setlength{\abovecaptionskip}{0cm}
  \renewcommand\arraystretch{1.3}
  \small
  \centering
  \caption{Ablation study on the embedding depth of M2DM and LEEM modules. The best results are highlighted in bold.}
  \label{tab:EmbedDepth}
  \setlength{\tabcolsep}{5pt}
    \begin{tabular}{c|c|cc}
      \toprule
      \multicolumn{2}{c|}{\textbf{Depth}} & \multicolumn{2}{c}{\textbf{Detection Performance}} \\
      M2DM-depth & LEEM-depth & AP\textsubscript{50} (\%) $\uparrow$ & Recall (\%) $\uparrow$ \\
      \midrule
      1 & \multirow{4}{*}{2} & 77.4 & 86.8 \\
      2 &                    & \textbf{77.9} & 87.2 \\
      3 &                    & 77.6 & \textbf{87.5} \\
      4 &                    & 77.5 & 87.2 \\
      \midrule
      \multirow{4}{*}{2} & 1 & 77.4 & 86.9 \\
                         & 2 & \textbf{77.9} & \textbf{87.2} \\
                         & 3 & 77.0 & 86.7 \\
                         & 4 & 77.2 & 86.8 \\
      \bottomrule
    \end{tabular}
\end{table}

The results demonstrate that embedding both M2DM and LEEM in the shallow layers (i.e., the first two stages) achieves the best detection performance, with the highest values in both AP\textsubscript{50} and Recall. This indicates that the rich edge and texture information in shallow features is more conducive to effective modeling by our dynamic modulation and edge enhancement modules. In contrast, embedding the modules into deeper layers, although providing more semantic information, leads to the loss of fine-grained details, and the presence of high-frequency noise hinders feature perception and model optimization. Consequently, we choose to introduce these two modules in the shallow stages of the backbone to fully exploit their capability in capturing fine-scale targets.

\subsubsection{Impact of the Backbone Depth}
As shown in Tab.~\ref{tab:BackboneDepth}, our proposed AuxDet consistently outperforms the baseline across different backbone depths, demonstrating strong generalization ability and adaptability to various network capacities. Notably, the best performance is achieved when using ResNet-50 as the backbone, which strikes a balance between semantic richness and spatial detail. Shallow backbones like ResNet-18 or ResNet-34 may lack sufficient semantic abstraction, while deeper ones such as ResNet-101, despite offering richer semantics, tend to dilute fine-grained details that are critical for small target detection.


\begin{table}[htbp]
  \centering
  \caption{Ablation study on the backbone depth. The best results are highlighted in bold.}
  \label{tab:BackboneDepth}
  \renewcommand{\arraystretch}{1.3} 
  \setlength{\tabcolsep}{6pt} 
  \footnotesize 
    \begin{tabular}{c|cc|cc}
      \toprule
      \multirow{2}{*}{\textbf{Depth}} & \multicolumn{2}{c|}{\textbf{Baseline}} & \multicolumn{2}{c}{\textbf{AuxDet}} \\
       & AP\textsubscript{50} (\%)$\uparrow$ & Recall (\%)$\uparrow$ & AP\textsubscript{50} (\%)$\uparrow$ & Recall (\%)$\uparrow$ \\
      \midrule
      18  & 72.7 & 83.4 & 74.1 & 85.2 \\
      34  & 74.2 & 84.6 & 75.7 & 85.8 \\
      \rowcolor[rgb]{0.9,0.9,0.9}
      \textbf{50}  & 74.7 & 84.5 & \textbf{77.9} & \textbf{87.2} \\
      101 & 74.6 & 84.4 & 76.4 & 86.2 \\
      \bottomrule
    \end{tabular}
\end{table}


Moreover, regardless of the backbone depth, AuxDet consistently brings significant improvements in both AP\textsubscript{50} and Recall, indicating its robustness and plug-and-play flexibility across different backbone architectures. This confirms that our framework can serve as an effective and scalable enhancement to existing detection pipelines.

\subsubsection{Impact of Metadata Composition}
As shown in Tab.~\ref{tab:MetaComp}, under controlled experimental conditions with consistent network architecture parameters, we systematically analyze the synergistic effects of multi-source metadata integration within the Multi-Modal Dynamic Modulation Module on detection performance. The results demonstrate the necessity of metadata fusion: incorporating any single metadata dimension (sensor platform, image resolution, or spectral band) improves average precision, while their combined input further significantly enhances detection effectiveness, validating the generalizable benefits of auxiliary metadata in facilitating robust cross-domain adaptation.

\begin{table}[htbp]
  \centering
  \caption{Ablation study on metadata composition. The best results are highlighted in bold.}
  \label{tab:MetaComp}
  \renewcommand{\arraystretch}{1.3} 
  \setlength{\tabcolsep}{6pt} 
  \footnotesize 
    \begin{tabular}{ccc|cc}
      \toprule
      \multicolumn{3}{c|}{\textbf{Metadata Components}} & \multicolumn{2}{c}{\textbf{Detection Performance}} \\
      Sensor & Resolution & Band & AP\textsubscript{50} (\%)$\uparrow$ & Recall (\%)$\uparrow$ \\
      \midrule
      &   &   & 74.7 & 84.5 \\
      \ding{51} &   &   & 76.7 & 86.3 \\
      & \ding{51} &   & 76.9 & 86.6 \\
      &   & \ding{51} & 76.5 & 86.2 \\
      \rowcolor[rgb]{0.9,0.9,0.9}
      \ding{51} & \ding{51} & \ding{51} & \textbf{77.2} & \textbf{86.7} \\
      \bottomrule
    \end{tabular}
\end{table}

%% file: journal-contents/8-conclusion.tex
\section{Conclusion} \label{sec:conclusion}
In this paper, we tackle the critical challenge of Omni-IRSTD by introducing a metadata-driven paradigm that redefines the traditional reliance on visual-only detection frameworks.
Through a high-dimensional fusion module, AuxDet dynamically integrates metadata semantics with visual features, effectively addressing domain heterogeneity by tailoring representations to the unique characteristics of each input sample.
Moreover, we propose the LEEM module that refines target details to enhance detection precision even amidst complex backgrounds.
Extensive experiments on the WideIRSTD-Full benchmark reveal that AuxDet significantly outperforms state-of-the-art approaches, affirming the indispensable role of auxiliary metadata in improving both robustness and accuracy across diverse omni-domain IRSTD scenarios.
\section*{Acknowledgment}

The authors would like to thank the editor and the anonymous reviewers for their critical and constructive comments and suggestions.
We acknowledge the Tianjin Key Laboratory of Visual Computing and Intelligent Perception (VCIP) for their essential resources. Computation is partially supported by the Supercomputing Center of Nankai University (NKSC).